\documentclass[sigconf]{acmart}
\pdfoutput=1
\usepackage{algorithm}
\usepackage{algpseudocode}
\usepackage{multirow}
\usepackage{mwe}
\usepackage{cuted}
\usepackage{caption}
\usepackage{stfloats}
\usepackage{graphicx}
\usepackage{makecell}
\usepackage{ragged2e}

\setcopyright{acmlicensed}
\copyrightyear{2026}
\acmYear{2026}

\acmConference[WWW '26]
  {The ACM Web Conference 2026}
  {April 13--17, 2026}
  {Dubai, United Arab Emirates}

\acmDOI{}
\acmISBN{}

\begin{document}

\title{TimeMAR: Multi-Scale Autoregressive Modeling for Unconditional Time Series Generation}


\author{Xiangyu Xu}
\authornote{Xiangyu Xu and Qingsong Zhong contributed equally to this work.}
\email{10225101535@stu.ecnu.edu.cn}
\affiliation{%
  \institution{Data Science and Engineering,\\  East China Normal University}
  \city{Shanghai}
  \country{China}
}

\author{Qingsong Zhong}
\authornotemark[1]
\email{xxrelax@stu.ecnu.edu.cn}
\affiliation{%
  \institution{Data Science and Engineering,\\  East China Normal University}
  \city{Shanghai}
  \country{China}
}

\author{Jilin Hu}
\authornote{Corresponding author.}
\email{hujilin@cs.aau.dk}
\affiliation{%
  \institution{Data Science and Engineering,\\  East China Normal University}
  \city{Shanghai}
  \country{China}
}
\affiliation{%
  \institution{Engineering Research Center of Blockchain Data Management (ECNU), Ministry of Education}
  \city{Shanghai}
  \country{China}
}


\begin{abstract}
Generative modeling offers a promising solution to data scarcity and privacy challenges in time series analysis. However, the structural complexity of time series, characterized by multi-scale temporal patterns and heterogeneous components, remains insufficiently addressed. In this work, we propose a structure-disentangled multiscale generation framework for time series. Our approach encodes sequences into discrete tokens at multiple temporal resolutions and performs autoregressive generation in a coarse-to-fine manner, thereby preserving hierarchical dependencies. To tackle structural heterogeneity, we introduce a dual-path VQ-VAE that disentangles trend and seasonal components, enabling the learning of semantically consistent latent representations. Additionally, we present a guidance-based reconstruction strategy, where coarse seasonal signals are utilized as priors to guide the reconstruction of fine-grained seasonal patterns. Experiments on six datasets show that our approach produces higher-quality time series than existing methods. Notably, our model achieves strong performance with a significantly reduced parameter count and exhibits superior capability in generating high-quality long-term sequences. Our implementation is available at \href{https://anonymous.4open.science/r/TimeMAR-BC5B}{https://anonymous.4open.science/r/TimeMAR-BC5B}.
\end{abstract}

\ccsdesc[300]{Computing methodologies~Machine learning}

\keywords{Time series generation, Autoregressive models, Multiscale modeling}

\maketitle

\section{Introduction}
Time-series data play a pivotal role in a wide range of high-impact domains, including finance~\cite{finance1, finance2, finance3, finance4, finance5, finance6}, healthcare~\cite{healthcare1, healthcare2, healthcare3, healthcare4}, and energy~\cite{energy1, energy2}, where they underpin critical decision making and intelligent systems~\cite{qiu2024tfb, qiu2025tab, qiu2025duet, qiu2025dag, liu2025rethinking, liu2025towards, timecma2025liu}. This role is amplified in modern Web, Mobile, and Web of Things (WoT) ecosystems, such as mobile health applications, federated sensor networks, and decentralized edge platforms, which generate temporal data at an unprecedented scale. While supervised learning is a powerful paradigm for leveraging such data, its reliance on large, centralized datasets stands in direct conflict with the inherent nature of these systems. Specifically, the core characteristics of Web/Mobile/WoT environments—including decentralization, resource-constrained devices, intermittent connectivity, and a strong emphasis on user privacy—create formidable barriers to data acquisition. For instance, medical records are protected by strict privacy regulations~\cite{shickel2017deep}, and financial data is typically withheld due to commercial sensitivity~\cite{bouchaud2018trades}, etc. Furthermore, the sheer cost and complexity of aggregating data from millions of distributed nodes is often prohibitive~\cite{miao2024less}. Consequently, these factors pose a critical bottleneck, severely hindering the scalability and applicability of supervised learning in real-world time-series scenarios.

\begin{figure}[t]
    \centering
    \includegraphics[width=0.5\textwidth]{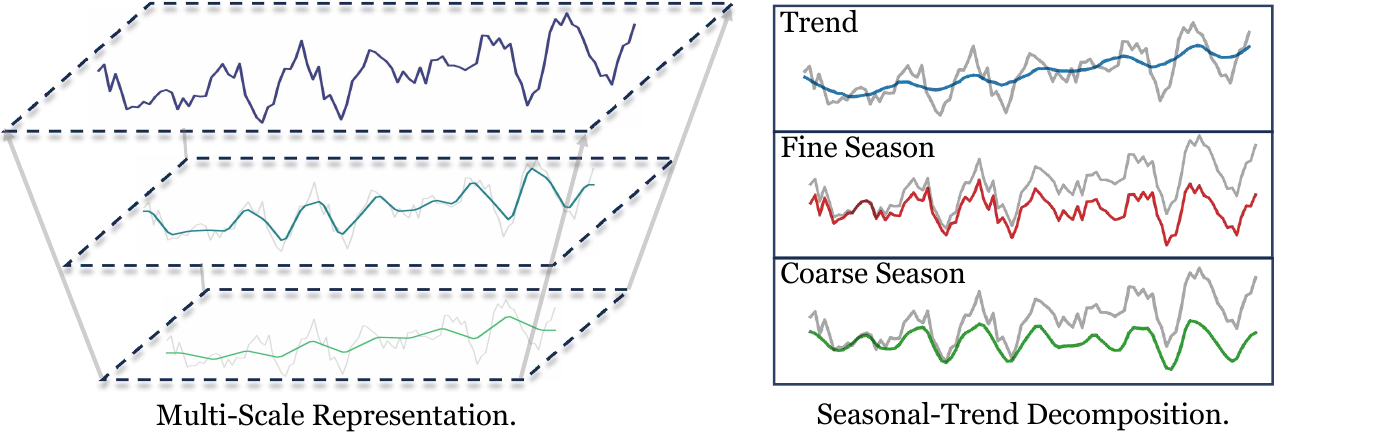}
        \captionof{figure}
        {Multi-Scale Representation of Time Series and Structural Decomposition. Left: Multi-Scale Representation. The series is further factorized across multiple resolutions, allowing hierarchical modeling. Right: Seasonal-Trend Decomposition. The input time series (gray) is decomposed into three interpretable components: underlying trend (top), fine-grained seasonality (middle), and coarse-grained seasonality (bottom).}
    \label{fig:multi_scale}
\end{figure}

Recently, generative modeling has emerged as a promising solution. It synthesizes realistic yet non-sensitive sequences, enabling downstream tasks without direct access to raw data. This privacy-preserving data generation mechanism can benefit large-scale web applications, enhance mobile user experience, and support intelligent decision-making in IoT ecosystems where data accessibility and privacy are critical concerns. Recent attempts of GANs~\cite{yoon2019time}, VAEs~\cite{desai2021timevae}, diffusion models~\cite{difftime,diffwave,diffusionts,timedit}, and autoregressive transformers~\cite{sdformer} for time-series generation have achieved encouraging results. However, existing methods struggle to address the structural complexities inherent in time series, particularly multi-scale temporal dependencies and structural heterogeneity. Specifically, many models operate at a single temporal resolution, limiting their ability to learn hierarchical patterns across different time scales. Furthermore, trend and seasonal components are often represented within a shared latent space or processed through a common encoder, which blurs their distinct temporal behaviors and impairs accurate generation.

\textbf{First}, real-world time series often exhibit inherently multi-scale structures. Temporal patterns manifest at different granularities—such as yearly, monthly, daily, or hourly—and these patterns are hierarchically organized. For instance, monthly electricity consumption reflects broad seasonal shifts, while daily usage patterns capture weekday-weekend cycles, and hourly data reveals intra-day fluctuations like morning and evening peaks. These scales are not isolated; finer patterns are often conditioned on coarser dynamics. Modeling such hierarchical dependencies requires mechanisms that explicitly represent and generate time series across multiple resolutions rather than relying on a single-scale representation, which may overlook important contextual dependencies.
\textbf{Second}, time-series are composed of structurally heterogeneous components with varying modeling difficulty. Trend components are usually smooth and low-frequency, easier to model, whereas seasonal components exhibit high-frequency, non-stationary fluctuations, sensitive to external factors. Traditional decomposition methods (e.g., STL) can separate these components, but reconstructing them faithfully and jointly within a generative framework remains challenging~\cite{qiu2025DBLoss}. These structural characteristics can be intuitively illustrated by the multi-scale representations  and decomposition shown in Figure~\ref{fig:multi_scale}.

To address these limitations, we propose a structure-disentangled multiscale time-series generation framework, combining coarse-to-fine autoregressive modeling, dual-path VQ-VAE encoding, and coarse-frequency guided decoding. Specifically, we first encode time-series into discrete tokens at multiple temporal resolutions and perform autoregressive generation from coarse to fine scales, explicitly preserving hierarchical consistency. To disentangle trend and seasonal structures, we design a dual-path VQ-VAE encoder that separately encodes the trend/coarse seasonal and fine seasonal components. This design enables the model to learn latent tokens with structural semantics. Furthermore, to improve generation quality for high-frequency components, we introduce a guidance-based reconstruction strategy: coarse seasonal representations are used as low-frequency priors to stabilize the generation of fine seasonal signals, inspired by multirate signal processing~\cite{vetterli1995wavelets}.

We evaluate our framework on six diverse datasets, including both real-world and synthetic time series. Experiments show that our method generates higher-quality time series than existing approaches. Notably, our approach achieves robust performance even with a significantly reduced parameter count and excels at generating high-quality long-term sequences.

Our main contributions are summarized as follows:
\begin{itemize}
    \item We propose a structure-disentangled multiscale time series generation framework that combines coarse-to-fine token modeling, dual-path encoding, and coarse-guided reconstruction.
    
    \item We design a dual-path VQ-VAE encoder to disentangle trend and seasonal components, enabling structurally aligned latent representations.
    
    \item We introduce a guidance-based reconstruction strategy that uses coarse seasonal priors to stabilize and refine the generation of high-frequency patterns.
    
    \item  We conduct comprehensive experiments on six real-world and synthetic datasets, demonstrating that our method delivers superior generation quality, robust performance with fewer parameters, and outstanding long-term sequence generation compared to existing methods.
\end{itemize}

\section{Related Work}
\label{gen_inst}

\subsection{Time Series Generation}
Early methods for time series generation predominantly relied on Generative Adversarial Networks (GANs)~\cite{GAN}. TimeGAN~\cite{yoon2019time} augments the adversarial loss with a supervised objective to better capture temporal dynamics and dependency structures, but still inherits the well-known issues of GANs—mode collapse and training instability. Subsequently, TimeVAE~\cite{desai2021timevae} introduced a structured Variational Autoencoder (VAE) that separates trend and seasonal components via dedicated priors, yielding high-fidelity reconstructions. Nonetheless, VAEs are prone to posterior collapse, limiting the generative diversity and quality.

Diffusion-based methods have recently emerged as a new paradigm for time series generation. In audio synthesis, DiffWave~\cite{diffwave} was the first to demonstrate the superior performance of Denoising Diffusion Probabilistic Models (DDPM) for high-fidelity waveform generation; DiffTime~\cite{difftime} employed guided diffusion to address constrained time series synthesis; Diffusion-TS\cite{diffusionts} integrated interleaved temporal representations into an encoder–decoder Transformer to produce high-quality multivariate time series. However, its reliance on lengthy multi-step denoising schedules results in extremely slow inference, which significantly limits its practicality. 

To strike a balance between efficiency and representational power, the recent SDformer~\cite{sdformer} leverages similarity-driven vector quantization to learn high-quality discrete token representations of time series, followed by discrete Transformer-based distribution modeling at the token level. This approach markedly outperforms existing continuous and diffusion-based methods. However, such performance gains come at the cost of a substantially increased parameter size, which raises concerns about scalability, especially in Web, mobile, and Web of Things (WoT) environments—where systems frequently handle very long sequential data. Moreover, the next-token prediction mechanism in SDformer is prone to error accumulation over long horizons, which can severely degrade global coherence in extended sequences and lead to significant deterioration in the quality of generated time series.

\subsection{Multi-Scale Modeling for Time Series}

Multi-scale feature extraction has widely been used for hierarchical feature extraction in fields such as computer vision ~\cite{muti-scale_cv1}\cite{muti-scale_cv2}\cite{muti-scale_cv3} and natural language processing~\cite{mutiscalenlp1}\cite{mutiscalenlp2}\cite{mutiscalenlp3}. More recently, multi-scale architectures have been adapted for time series forecasting, capturing temporal patterns at different resolutions. N-HiTS\cite{challu2023nhits} employs hierarchical interpolation and multi-rate data sampling to model hierarchical features of time series. Pyraformer\cite{liu2022pyraformer} uses pyramidal attention to summarize features across scales. Scaleformer\cite{shabani2022scaleformer} incrementally refines the predictions across scales, enhancing scale awareness. More recently, by introducing patch division with multiple patch sizes, Pathformer\cite{chen2024pathformer} models multi-scale characteristics via patch divisions of varying sizes.  TimeMixer++\cite{wang2024timemixer++} and TimeMixer\cite{wang2024timemixer} present a multiscale mixing architecture that leverages disentangled series and enables hierarchical interaction across different scales and resolutions. For anomaly detection, MODEM\cite{zhong2025multiresolution} employs a coarse-to-fine diffusion paradigm effectively capturing the non-stationarity to better differentiate between anomalies and normal patterns. While most of these approaches are designed for forecasting or detection, their success demonstrates the importance of multi-scale temporal representations. However, multi-scale mechanisms remain underexplored in the generative domain.

\begin{figure}[t]
  \centering
  \begin{minipage}[b]{0.40\textwidth}
    \centering
    \includegraphics[width=0.93\textwidth]{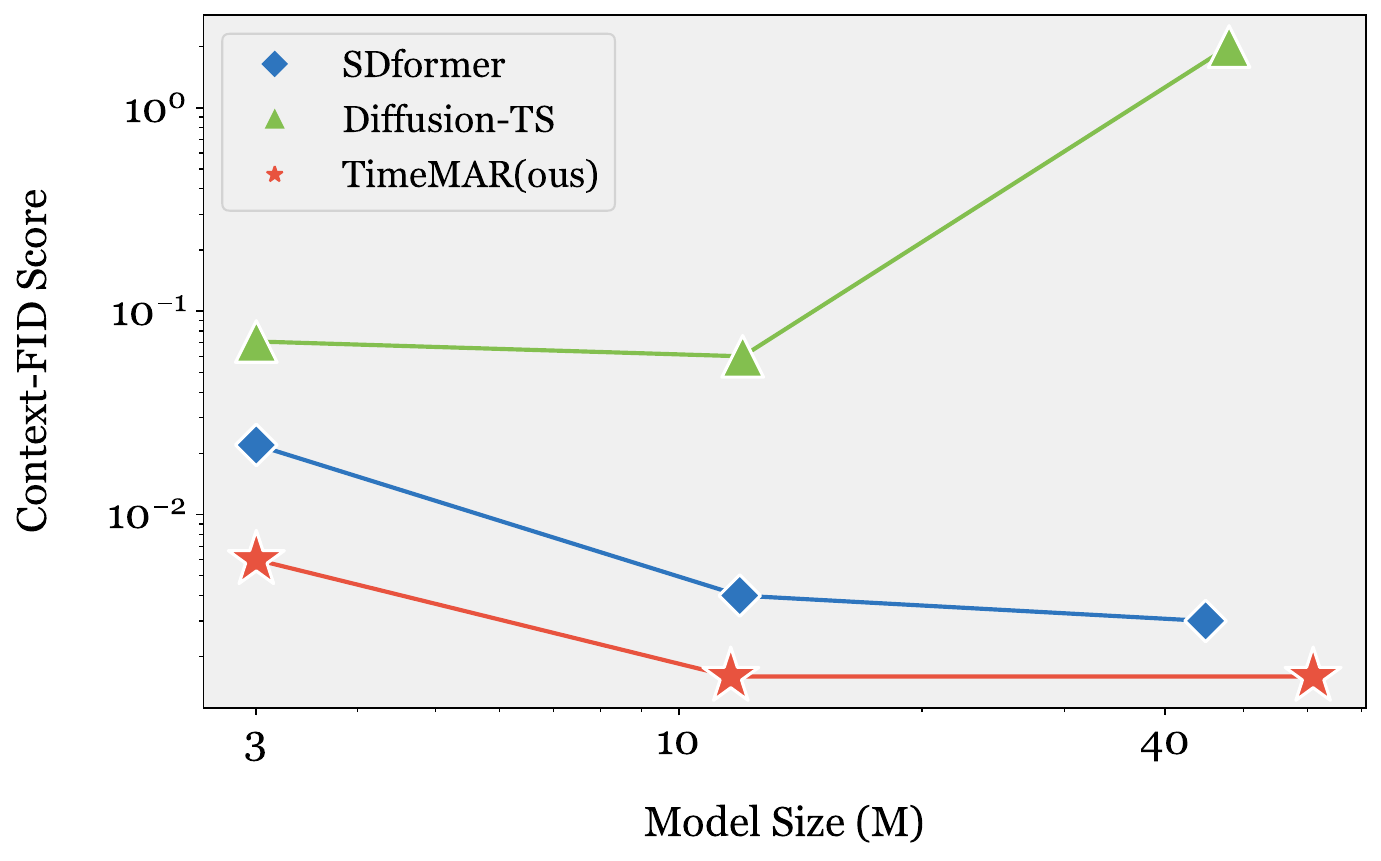}
    \caption{Scaling behavior of different model families on the Context-FID metric on Energy.}
    \label{fig:fig1}
  \end{minipage}
\end{figure}

\subsection{Autoregressive Models}

Autoregressive modeling has been successfully extended beyond language to various high-dimensional modalities. PixelCNN~\cite{pixelcnn} pioneered this shift by factorizing images pixel-by-pixel, modeling each pixel conditioned on its causal neighborhood. iGPT\cite{igpt} was the first to adopt a GPT-2–sized autoregressive model for learning powerful image representations. Building on VQVAE\cite{vqvae}, VQGAN\cite{vqgan} quantizes image patches into discrete tokens and employs a decoder-only Transformer to predict these tokens, enabling high-fidelity, high-resolution synthesis. MaskGIT~\cite{maskgit} introduces a mask-and-refine strategy that predicts large subsets of tokens in parallel, dramatically reducing generation steps. RQ-VAE\cite{rqvae} incorporated residual quantization to reduce quantization error; VAR\cite{Var} proposed a multi-scale autoregressive framework with a "next-scale prediction" mechanism, marking the first time a GPT-style autoregressive model surpassed leading diffusion models in image synthesis quality. SAR3D\cite{sar3d} successfully adapts this next-scale prediction strategy to 3D object generation, attesting to its modality-agnostic potential.

\begin{figure*}[h]
    \centering
    \includegraphics[width=0.9\textwidth]{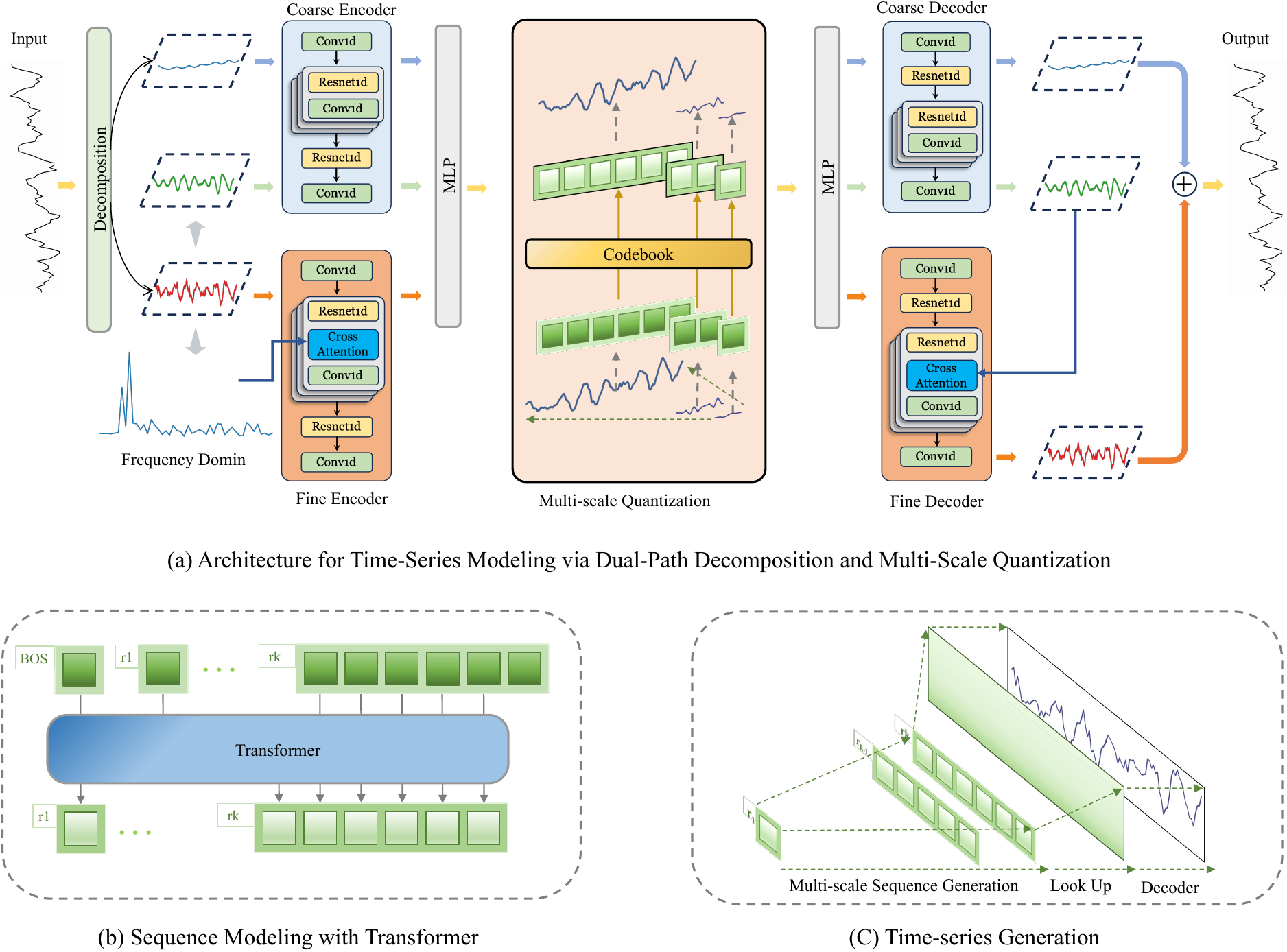}
    \captionof{figure}{
        Overview of the proposed TimeMAR framework. 
        (a) Architecture of the Multi-Scale VQ-VAE Module, which decomposes the input sequence and encodes components into multi-scale discrete representations using coarse and fine encoders. 
        (b) Sequence Modeling with Transformer, where a GPT-style autoregressive model learns token dependencies. 
        (c) Time-Series Reconstruction from Tokens, where generated tokens are decoded into continuous sequences.
    }
    \label{fig:framework}
\end{figure*}

\section{Preliminaries}
\subsection{Definitions}

\textit{Definition 1 (Multivariate time series)}.  
A multivariate time series of length $\tau$ is a sequence  
$
\mathbf{X}_{1:\tau} = (\mathbf{x}_1, \ldots, \mathbf{x}_\tau) \in \mathbb{R}^{\tau \times d},
$ 
where each $\mathbf{x}_t \in \mathbb{R}^d$ denotes the $d$-dimensional observation at time step $t$.  
If $d=1$, the series is called \emph{univariate}; if $d>1$, it is called \emph{multivariate}.  
We use $\mathbf{x}_{t,j} \in \mathbb{R}$ to denote the $j$-th dimension of $\mathbf{x}_t$, and $\mathbf{X}_{:,j} \in \mathbb{R}^\tau$ to denote the full trajectory of the $j$-th dimension over time.

\medskip

\textit{Definition 2 (Multi-scale temporal patterns)}.  
A time series often exhibits dynamics at multiple temporal resolutions.  
Formally, a sequence $\mathbf{X}_{1:\tau}$ can be decomposed into components $\{\mathbf{X}^{(s)}_{1:\tau_s}\}_{s=1}^S$, where each $\mathbf{X}^{(s)}$ captures temporal patterns at scale $s$ with resolution $\tau_s$.  
Modeling these multi-scale dependencies is critical for preserving both long-term trends and short-term variations in generative tasks.

\subsection{Problem Statement}
We consider the task of unconditional generation of multivariate time series. Given a dataset $\mathcal{D} = \{\mathbf{X}_{1:\tau}^{(i)}\}_{i=1}^{n}$ containing $n$ independent sequences, our objective is to learn a generative model $p_\theta(\mathbf{X}_{1:\tau})$ that closely approximates the underlying data distribution $p_{\text{data}}(\mathbf{X}_{1:\tau})$. This model should enable sampling of new sequences $\hat{\mathbf{X}}_{1:\tau} \sim p_\theta(\cdot)$ that exhibit similar statistical and temporal characteristics to those in the real dataset.

\section{Methods}
\subsection{Overall Framework}


To achieve high-quality time series generation, we propose \textbf{Time\-MAR}, 
a two-stage generative framework illustrated in Figure~\ref{fig:framework}. 
TimeMAR is built upon three core components:

\begin{itemize}
    \item \textbf{Trend–Seasonal Decomposition Module}: 
    explicitly disentangles complex temporal patterns.
    
    \item \textbf{Multi-Scale Vector-Quantized Autoencoder (VQ-VAE)}: 
    encodes the decomposed signals into discrete tokens across multiple resolutions.
    
    \item \textbf{Transformer-Based Autoregressive Module}: 
    models hierarchical temporal dependencies among the discrete tokens.
\end{itemize}

The framework operates in two stages:
\begin{enumerate}
    \item \textbf{Tokenization}: 
    the input series is decomposed into trend and seasonal components, 
    which are further encoded into discrete multi-scale latent tokens.
    \item \textbf{Sequence Modeling}: 
    a Transformer-based autoregressive module captures hierarchical and temporal dependencies 
    among the tokens. During generation, the module autoregressively produces token sequences, 
    which are then decoded back into continuous time series.
\end{enumerate}

In the following sections, we describe each module in detail.

\subsection{The Trend-Seasonal Decomposition Module}
In practice, diverse domains provide clear evidence of trend–seasonal mixtures: 
financial markets blend long-term growth with intraday volatility, 
Mujoco control trajectories combine long-horizon motion trends with rapid oscillatory behaviors, 
and energy usage reflects long-term demand shifts overlaid with daily patterns. 
Real-world time series therefore often exhibit complex dynamics, mixing slowly evolving trends with rapidly fluctuating seasonal patterns. 
Direct modeling of such mixtures complicates representation learning and limits generative performance. 
To address this issue, we propose a dedicated decomposition module that explicitly separates trend and seasonal components, 
enabling specialized modeling for each.

\textbf{Decomposition via Mixture-of-Experts (MoE).} Following FEDformer~\cite{FEDformer}, we adopt a MoE strategy to effectively extract the underlying trend from the input time series. Specifically, given an input multivariate series $\mathbf{X}\in\mathbb{R}^{T\times d}$, we first generate multiple candidate trend signals at various temporal resolutions using average pooling filters with different kernel sizes. These candidates are then dynamically aggregated using data-dependent mixing weights. Formally, this decomposition process is defined as:

\begin{align*}
\mathbf{X}_t &= \text{Softmax}(L(\mathbf{X})) \cdot F(\mathbf{X}) , \\
\mathbf{X}_s &= \mathbf{X} - \mathbf{X}_t,
\end{align*}
where $F(\cdot)$ denotes average filters and $L(\cdot)$ produces their mixing weights. As a result, we obtain the trend component $\mathbf{X}_t$, which captures smooth low-frequency dynamics, and the residual component $\mathbf{X}_s$, which retains high-frequency seasonal variations.

\textbf{Guidance-based Reconstruction Strategy.}
The seasonal signal $\mathbf{X}_s$ is significantly more challenging to reconstruct due to its non-stationarity and rapid fluctuations. To enhance reconstruction stability, we propose a guidance-based reconstruction strategy inspired by multirate signal processing~\cite{vetterli1995wavelets}. The key idea is to extract a coarse seasonal approximation that preserves dominant low-frequency content, which serves as conditioning input during reconstruction.
This is implemented via a downsampling-then-upsampling operation applied to the seasonal component:
\[
\mathbf{X}_c = \text{Upsample}(\text{Downsample}(\mathbf{X}_s)) ,
\]
where $\mathbf{X}_c$ serves as a coarse and more stable version of the seasonal pattern. During decoding, $\mathbf{X}_c$ is incorporated into the fine seasonal reconstruction via a cross-attention mechanism, guiding the decoder to focus on meaningful low-frequency priors while refining local variations.

\subsection{Multi-Scale Reconstruction}
To effectively capture the hierarchical temporal structure of time series, we employ a multi-scale Vector Quantized Variational Autoencoder (VQ-VAE). This module discretizes the latent representations of decomposed components (trend, coarse seasonal, and fine seasonal) across multiple resolutions, enabling coarse-to-fine generation and structurally aligned decoding. The process consists of three parts: (i) dual-path latent encoding, (ii) hierarchical multi-scale quantization, and (iii) Guidance-based reconstruction.

\textbf{Dual-path Latent Encoding.}
We design two distinct encoders to model different temporal characteristics. The trend $\mathbf{X}_t$ and coarse seasonal $\mathbf{X}c$ components, which are low-frequency and smooth, are encoded by a shared coarse encoder $\mathrm{Encoder}_{\text{coarse}}$. In contrast, the fine seasonal component $\mathbf{X}s$, which contains high-frequency and non-stationary variations, is encoded via a dedicated high-resolution encoder $\mathrm{Encoder}_{\text{fine}}$.

To better capture the oscillatory and high-frequency nature of the fine seasonal component, $\mathrm{Encoder}_{\text{fine}}$ incorporates a frequency-aware enhancement module. We apply the Fast Fourier Transform (FFT) to $\mathbf{X}s$ to obtain its frequency-domain representation. For each frequency bin, we extract both its amplitude and phase, concatenate them, and pass the result through a multilayer perceptron (MLP) to obtain a compact spectral embedding $\mathbf{H}{\mathrm{freq}}$. These are fused with time-domain features $\mathbf{H}_{\mathrm{time}}$ from a convolutional backbone via cross-attention:
\[
\mathbf{Z} = \sigma\left(\frac{\mathcal{P}_t(\mathbf{H}_{\mathrm{time}})\mathcal{P}_f(\mathbf{H}_{\mathrm{freq}})^\top}{\sqrt{d_k}}\right)\mathcal{P}_v(\mathbf{H}_{\mathrm{freq}}) ,
\]
where $\mathcal{P}_t$, $\mathcal{P}_f$, and $\mathcal{P}_v$ are linear projections for the queries, keys, and values, respectively, and $d_k$ denotes the key dimensionality.

The latent features are then computed as:
\begin{align*}
\mathbf{z}_t &= \mathrm{Encoder}_{\text{coarse}}(\mathbf{X}_t) , \\
\mathbf{z}_c &= \mathrm{Encoder}_{\text{coarse}}(\mathbf{X}_c) ,\\
\mathbf{z}_s &= \mathrm{Encoder}_{\text{fine}}(\mathbf{X}_s) .
\end{align*}

\textbf{Hierarchical Multi-Scale Quantization.}
We aggregate the encoded features into a unified latent representation using an MLP:
\[
\mathbf{z}_{\mathrm{fus}} = f_{\mathrm{fuse}}([\mathbf{z}_t, \mathbf{z}_c, \mathbf{z}_s]).
\]
The fused latent is reshaped into multiple temporal resolutions (e.g., $1 \times 1$, $1 \times 2$, $\ldots$, $1 \times K$). Each resolution is quantized via a shared codebook, yielding $K$ multi-scale discrete token maps $\{r_1, r_2, \ldots, r_K\}$, which collectively constitute the discrete token sequence $R$.
To preserve fine-grained detail, residuals between quantized and original representations are propagated across scales. This coarse-to-fine tokenization enables early capture of global structures and progressive refinement of local variations.

\textbf{Guidance-Based Reconstruction.}To reconstruct the original components, we first interpolate and de-embed the quantized token sequence:
\[
\widehat{\mathbf{z}}_{\mathrm{fus}} = \sum_{k=1}^K \operatorname{Interpolate}(\mathrm{Embed}^{k}(r_k)).
\]
We then recover individual components via an MLP:
\renewcommand{\arraystretch}{1.2} 
\[
\begin{bmatrix}
\widehat{\mathbf{z}}_t \\
\widehat{\mathbf{z}}_c \\
\widehat{\mathbf{z}}_s
\end{bmatrix}
= f_{\mathrm{split}}(\widehat{\mathbf{z}}_{\mathrm{fus}}).
\]

The trend and coarse seasonal components are decoded using the shared coarse decoder $\mathrm{Decoder}_{\text{coarse}}$, while the fine seasonal component is reconstructed via $\mathrm{Decoder}_{\text{fine}}$, guided by the coarse seasonal output via cross-attention:
\begin{align*}
\widehat{\mathbf{T}} &= \mathrm{Decoder}_{\text{coarse}}(\widehat{\mathbf{z}}_t) , \\
\widehat{\mathbf{C}} &= \mathrm{Decoder}_{\text{coarse}}(\widehat{\mathbf{z}}_c) , \\
\widehat{\mathbf{S}} &= \mathrm{Decoder}_{\text{fine}}(\widehat{\mathbf{z}}_s, \mathrm{Attention}(\widehat{\mathbf{z}}_s, \widehat{\mathbf{C}})) .
\end{align*}

The final output is:
\[
\widehat{\mathbf{X}} = \widehat{\mathbf{T}} + \widehat{\mathbf{S}} .
\]

\subsection{Multi-Scale Reconstruction Objective}

The training objective of the multi-scale reconstruction integrates multiple loss terms to achieve accurate reconstruction and effective disentanglement of temporal patterns. The composite loss is defined as:

\begin{equation*}
\begin{split}
\mathcal{L} = & \ \lambda_{\text{rec}} \mathcal{L}_{\text{rec}} 
+ \lambda_{\text{vq}} \mathcal{L}_{\text{vq}} 
+ \lambda_{\text{trend}} \mathcal{L}_{\text{rec\_trend}} \\
& + \lambda_{\text{season}} \mathcal{L}_{\text{rec\_season}} 
+ \lambda_{\text{fourier}} \mathcal{L}_{\text{fourier}} 
+ \lambda_{\text{coarse}} \mathcal{L}_{\text{rec\_coarse}} .
\end{split}
\end{equation*}

\begin{table*}[t]
    \centering
    \scriptsize
    \setlength{\tabcolsep}{4pt}
    \renewcommand{\arraystretch}{1.10}
    {\fontsize{8.5}{9.5}\selectfont
    \begin{tabular}{lllcccccc}
    \toprule
    \textbf{Metric} & \textbf{Model Size} & \textbf{Method} & \textbf{Sines} & \textbf{Stocks} & \textbf{ETTh} & \textbf{MuJoCo} & \textbf{Energy} & \textbf{fMRI} \\
    \midrule
    
    \multirow{10}{*}{\shortstack[c]{\textbf{Discriminative}\\\textbf{Score} $\downarrow$}}
    & \multirow{2}{*}{Large(>40M)} 
      & \textbf{TimeMAR-L} & \textbf{0.003$\pm$.001} & \textbf{0.005$\pm$.003} & \textbf{0.003$\pm$.002} & \underline{0.006$\pm$.003} & \textbf{0.003$\pm$.002} & \textbf{0.013$\pm$.005} \\
    &  
      & SDformer     & \underline{0.006$\pm$.004} &  0.010$\pm$.006 &  \textbf{0.003$\pm$.001} &  0.008$\pm$.005 & 0.006$\pm$.004 &  \underline{0.017$\pm$.007} \\
    \cmidrule{2-9}
    & Small(<6M)
      & \textbf{TimeMAR-S}    & \textbf{0.003$\pm$.002} & \underline{0.007$\pm$.003} & \underline{0.004$\pm$.002} & \textbf{0.005$\pm$.003} & \underline{0.005$\pm$.002} & 0.114$\pm$.034 \\
    & 
      & Diffusion-TS & \underline{0.006$\pm$.007} & 0.067$\pm$.015 & 0.061$\pm$.009 & 0.008$\pm$.002 & 0.122$\pm$.003 & 0.167$\pm$.023 \\
    & 
      & TimeGAN     & 0.011$\pm$.008 & 0.102$\pm$.021 & 0.114$\pm$.055 & 0.238$\pm$.068 & 0.236$\pm$.012 & 0.484$\pm$.042 \\
    & 
      & TimeVAE     & 0.041$\pm$.044 & 0.145$\pm$.120 & 0.209$\pm$.058 & 0.230$\pm$.102 & 0.499$\pm$.000 & 0.476$\pm$.044 \\
    & 
      & Diffwave    & 0.017$\pm$.008 & 0.232$\pm$.061 & 0.190$\pm$.008 & 0.203$\pm$.096 & 0.493$\pm$.004 & 0.402$\pm$.029 \\
    & 
      & DiffTime    & 0.013$\pm$.006 & 0.097$\pm$.016 & 0.100$\pm$.007 & 0.154$\pm$.045 & 0.445$\pm$.004 & 0.245$\pm$.051 \\
    & 
      & Cot-GAN     & 0.254$\pm$.137 & 0.230$\pm$.016 & 0.325$\pm$.099 & 0.426$\pm$.022 & 0.498$\pm$.002 & 0.492$\pm$.018 \\
    \midrule
    
    \multirow{10}{*}{\shortstack[c]{\textbf{Context-FID}\\\textbf{Score} $\downarrow$}}
    & \multirow{2}{*}{Large(>40M)}
      & \textbf{TimeMAR-L} & \textbf{0.001$\pm$.000} & \textbf{0.001$\pm$.000} & \textbf{0.001$\pm$.000} & \textbf{0.003$\pm$.000} & \textbf{0.002$\pm$.000} & \textbf{0.008$\pm$.000} \\
    & 
      & SDformer & \textbf{0.001$\pm$.000} & \underline{0.002$\pm$.000} & 0.008$\pm$.001 & \underline{0.005$\pm$.001} & \underline{0.003$\pm$.000} & \underline{0.015$\pm$.001} \\
    \cmidrule{2-9}
    & Small(<6M)
      & \textbf{TimeMAR-S} & \textbf{0.001$\pm$.000} & \textbf{0.001$\pm$.000} & \underline{0.002$\pm$.000} & \textbf{0.003$\pm$.000} & 0.004$\pm$.000 & 0.176$\pm$.011 \\
    & 
      & Diffusion-TS & \underline{0.006$\pm$.000} & 0.147$\pm$.025 & 0.116$\pm$.010 & 0.013$\pm$.001 & 0.089$\pm$.024 & 0.105$\pm$.006 \\
    & 
      & TimeGAN     & 0.101$\pm$.014 & 0.103$\pm$.013 & 0.300$\pm$.013 & 0.563$\pm$.052 & 0.767$\pm$.103 & 1.292$\pm$.218 \\
    & 
      & TimeVAE     & 0.307$\pm$.060 & 0.215$\pm$.035 & 0.805$\pm$.186 & 0.251$\pm$.015 & 1.631$\pm$.142 & 14.449$\pm$.969 \\
    & 
      & Diffwave    & 0.014$\pm$.002 & 0.232$\pm$.032 & 0.873$\pm$.061 & 0.393$\pm$.041 & 1.031$\pm$.131 & 0.244$\pm$.018 \\
    & 
      & DiffTime    & 0.006$\pm$.001 & 0.236$\pm$.074 & 0.299$\pm$.044 & 0.188$\pm$.028 & 0.279$\pm$.045 & 0.340$\pm$.015 \\
    & 
      & Cot-GAN     & 1.337$\pm$.068 & 0.408$\pm$.086 & 0.980$\pm$.071 & 1.094$\pm$.079 & 1.039$\pm$.028 & 7.813$\pm$.550 \\

    \bottomrule
    \end{tabular}
    }
        \begin{flushleft}
    \end{flushleft}
    \caption{Generation performance on 24-length time series across all datasets. Lower values indicate better performance (↓).}
    \label{tab:evaluation-metrics}
\end{table*}

\textbf{Global Reconstruction Loss.} The overall reconstruction loss ensures end-to-end fidelity:
\[
\mathcal{L}_{\text{rec}} = \|\mathbf{X} - \widehat{\mathbf{X}}\|^2 .
\]

\textbf{Component-specific Reconstruction Losses.} To encourage disentanglement and specialized modeling, we introduce reconstruction terms for individual components:
\begin{align*}
\mathcal{L}_{\text{rec\_trend}} &= \|\mathbf{X}_t - \widehat{\mathbf{T}}\|^2 , \\
\mathcal{L}_{\text{rec\_season}} &= \|\mathbf{X}_s - \widehat{\mathbf{S}}\|^2 , \\
\mathcal{L}_{\text{rec\_coarse}} &= \|\mathbf{X}_c - \widehat{\mathbf{C}}\|^2 ,
\end{align*}
where $\mathbf{X}_t$, $\mathbf{X}_s$, and $\mathbf{X}_c$ denote the ground-truth trend, seasonal, and coarse components, while $\widehat{\mathbf{T}}, \widehat{\mathbf{S}}, \widehat{\mathbf{C}}$ are their reconstructions.

\textbf{Vector Quantization Loss.} Following standard VQ-VAE training, we employ:
\[
\mathcal{L}_{\text{vq}} = \|\text{sg}[\mathbf{z}_{\mathrm{fus}}] - \mathbf{e}\|^2 + \beta \|\mathbf{z}_{\mathrm{fus}} - \text{sg}[\mathbf{e}]\|^2 ,
\]
where $\text{sg}[\cdot]$ denotes the stop-gradient operation and $\beta$ is the commitment weight.

\textbf{Frequency-Domain Loss.} Inspired by Diffusion-TS, we adopt the Mean Absolute Error in the Fourier domain to better preserve periodic patterns, particularly within the seasonal component:
\[
\mathcal{L}_{\text{fourier}} = \|\mathcal{F}(\mathbf{X}_s) - \mathcal{F}(\widehat{\mathbf{S}})\|_1 ,
\]
where $\mathcal{F}(\cdot)$ denotes the Fast Fourier Transform.


\subsection{Autoregressive Modeling over Multi-Scale Discrete Tokens}

To model temporal dependencies across scales, we adopt a GPT-style Transformer decoder over the sequence of quantized tokens, as illustrated in Figure~\ref{fig:framework}(b). The token vocabulary is defined as:

\[
V = \{0, 1, \dots, N-1\} \cup \{[\mathrm{BOS}]\},
\]
where $N$ is the codebook size, and $[\mathrm{BOS}]$ denotes a special start-of-sequence token.

\paragraph{Training Setup.}
Given a ground-truth token sequence $(y_1, y_2, \dots, y_\ell)$, we construct the model input as
\[
\mathbf{y}_{\mathrm{in}} = ([\mathrm{BOS}], y_1, \dots, y_{\ell-1}).
\]

\paragraph{Objective Function.}
The model is optimized by minimizing the negative log-likelihood:
\[
\mathcal{L}(\theta) = -\sum_{i=1}^{\ell} \log P_\theta(y_i \mid y_0, y_1, \dots, y_{i-1}), 
\quad y_0 = [\mathrm{BOS}].
\]

\paragraph{Generation.}
During inference, the model starts with only the special token $[\mathrm{BOS}]$ as input, and then generates new tokens autoregressively:
\[
\hat{y}_i \sim P_\theta(y_i \mid \hat{y}_0, \dots, \hat{y}_{i-1}).
\]

This autoregressive framework captures fine-grained temporal dependencies within each resolution while also modeling hierarchical structures across different temporal scales.

\section{Experiments}

\subsection{Experimental Setup}

\textbf{Datasets.} 
We evaluate our model on six multivariate time series datasets, including four real-world benchmarks: Stocks, ETTh~\cite{zhou2021informer}, Energy, and fMRI; and two synthetic datasets: Sines~\cite{yoon2019time} and MuJoCo~\cite{mujoco}.
These datasets exhibit diverse temporal patterns, sampling resolutions, and structural characteristics, enabling a comprehensive assessment of model generalization and robustness.

\textbf{Metrics.}
We evaluate generation quality with three standard metrics: Discriminative Score~\cite{yoon2019time}, and Context-FID~\cite{fid}; their detailed definitions are given in Appendix D.2.

\textbf{Baselines.} 
We compare our method against representative generative models: TimeGAN (GAN-based), TimeVAE (VAE-based), DiffWave, DiffTime, and Diffusion-TS (diffusion-based), and SDformer (autoregressive transformer). These methods collectively span a wide range of generative paradigms and modeling assumptions, providing strong baselines for comparative evaluation.

\begin{table*}[ht]
\centering
    \scriptsize
    \setlength{\tabcolsep}{4pt}
    \renewcommand{\arraystretch}{1.10}
    {\fontsize{8.5}{9.5}\selectfont
    \begin{tabular}{lllcccccc}
    \toprule
    \textbf{Metric} & \textbf{Model Size} & \textbf{Method} & \textbf{ETTh-64} & \textbf{ETTh-128} & \textbf{ETTh-256} & \textbf{Energy-64} & \textbf{Energy-128} & \textbf{Energy-256} \\
    \midrule
    \multirow{10}{*}{\shortstack[c]{\textbf{Discriminative}\\\textbf{Score} $\downarrow$}} 
    & \multirow{2}{*}{Large(>40M)} 
      & \textbf{TimeMAR-L} & \textbf{0.003$\pm$.003} & \underline{0.008$\pm$.000} & \underline{0.006$\pm$.004} & \underline{0.006$\pm$.004} & \textbf{0.005$\pm$.004} & \textbf{0.007$\pm$.004} \\
    & 
      & SDformer    & {0.018$\pm$.007} & {0.013$\pm$.005} & {0.008$\pm$.006} & {0.010$\pm$.007} & \underline{0.013$\pm$.007} & \underline{0.017$\pm$.003} \\
    \cmidrule{2-9}
    & Small(<6M)
      & \textbf{TimeMAR-S} & \underline{0.004$\pm$.003} & \textbf{0.004$\pm$.003} & \textbf{0.004$\pm$.003} & \textbf{0.004$\pm$.002} & 0.015$\pm$.005 & 0.088$\pm$.012 \\
    &
      & Diffusion-TS & 0.106$\pm$.048 & 0.144$\pm$.060 & 0.060$\pm$.030 & 0.078$\pm$.021 & 0.143$\pm$.075 & 0.290$\pm$.123 \\
    & 
      & TimeGAN     & 0.227$\pm$.078 & 0.188$\pm$.074 & 0.442$\pm$.056 & 0.498$\pm$.001 & 0.499$\pm$.001 & 0.499$\pm$.000 \\
    & 
      & TimeVAE     & 0.171$\pm$.142 & 0.154$\pm$.087 & 0.178$\pm$.076 & 0.499$\pm$.000 & 0.499$\pm$.000 & 0.499$\pm$.000 \\
    & 
      & Diffwave    & 0.254$\pm$.074 & 0.274$\pm$.047 & 0.304$\pm$.068 & 0.497$\pm$.004 & 0.499$\pm$.001 & 0.499$\pm$.000 \\
    & 
      & DiffTime    & 0.150$\pm$.003 & 0.176$\pm$.015 & 0.243$\pm$.005 & 0.328$\pm$.031 & 0.396$\pm$.024 & 0.437$\pm$.095 \\
    & 
      & Cot-GAN     & 0.296$\pm$.348 & 0.451$\pm$.080 & 0.461$\pm$.010 & 0.499$\pm$.001 & 0.499$\pm$.001 & 0.498$\pm$.004 \\
    \midrule
    \multirow{10}{*}{\shortstack[c]{\textbf{Context-FID}\\\textbf{Score} $\downarrow$}} 
    & \multirow{2}{*}{Large(>40M)}
      & \textbf{TimeMAR-L} & \textbf{0.001$\pm$.000} & \textbf{0.002$\pm$.000} & \textbf{0.002$\pm$.000} & \textbf{0.002$\pm$.000} & \textbf{0.003$\pm$.000} & \textbf{0.008$\pm$.001} \\
    &
      & SDformer & {0.018$\pm$.003} & {0.024$\pm$.001} & {0.021$\pm$.001} & {0.031$\pm$.002} & {0.036$\pm$.002} & {0.041$\pm$.003} \\
    \cmidrule{2-9}
    & Small(<6M)
      & \textbf{TimeMAR-S} & \underline{0.005$\pm$.000} & \underline{0.003$\pm$.000} & \underline{0.003$\pm$.000} & \underline{0.005$\pm$.000} & \underline{0.006$\pm$.000} & \underline{0.013$\pm$.002} \\
    &
      & Diffusion-TS & 0.631$\pm$.058 & 0.787$\pm$.062 & 0.812$\pm$.038 & 0.135$\pm$.017 & 0.087$\pm$.019 & 0.126$\pm$.024 \\
    & 
      & TimeGAN     & 1.130$\pm$.102 & 1.553$\pm$.169 & 1.587$\pm$.128 & 1.230$\pm$.070 & 2.535$\pm$.372 & 5.032$\pm$.831 \\
    & 
      & TimeVAE     & 0.827$\pm$.146 & 1.062$\pm$.134 & 0.826$\pm$.093 & 2.662$\pm$.087 & 3.125$\pm$.106 & 3.768$\pm$.998 \\
    & 
      & Diffwave    & 1.543$\pm$.153 & 2.354$\pm$.170 & 2.899$\pm$.289 & 2.697$\pm$.418 & 5.552$\pm$.528 & 5.572$\pm$.584 \\
    & 
      & DiffTime    & 1.279$\pm$.083 & 2.554$\pm$.318 & 3.524$\pm$.830 & 0.762$\pm$.157 & 1.344$\pm$.131 & 4.735$\pm$.729 \\
    & 
      & Cot-GAN     & 3.008$\pm$.277 & 2.639$\pm$.427 & 4.075$\pm$.894 & 1.824$\pm$.144 & 1.822$\pm$.271 & 2.533$\pm$.467 \\
    \bottomrule
    \end{tabular}
    }
    \caption{Generation performance on ETTh and Energy datasets under varying input lengths (64/128/256). Lower values indicate better performance (↓).}
    \label{tab:scalable-input-length}
\end{table*}

\subsection{Main Results}

Table~\ref{tab:evaluation-metrics} presents the performance of TimeMAR in generating 24-length time series across multiple benchmark datasets. The results demonstrate that TimeMAR achieves highly competitive performance on all evaluation metrics. Notably, TimeMAR-L attains the lowest Discriminative Score on every dataset, indicating superior sample realism. When compared to the second-best model, SDformer, TimeMAR-L exhibits substantial relative improvements reducing the Discriminative Score by 50\% on Sines, Stocks, and Energy, by 25\% on MuJoCo, and by 23.5\% on fMRI. On the Context-FID metric, which reflects distributional and contextual fidelity, TimeMAR-L substantially outperforms all baseline methods. Specifically, it reduces Context-FID by 50.0\% on Stocks, 87.5\% on ETTh, 40.0\% on MuJoCo, 33.3\% on Energy, and 46.7\% on fMRI.

Furthermore, the large-scale variant of our model (TimeMAR-L) consistently ranks first across all metrics and datasets, while the small-scale variant (TimeMAR-S) significantly outperforms all other models in its parameter class (<6M). Notably, TimeMAR-S achieves highly competitive performance that is even comparable to large-scale models (>40M) on several datasets (e.g., Sines, Stocks, ETTh, and MuJo Co), which underscores its exceptional parameter efficiency.

To evaluate scalability, Table~\ref{tab:scalable-input-length} presents results for longer sequences (lengths 64, 128, and 256) on ETTh and Energy. Compared with the 24-length setting, TimeMAR-L maintains consistently strong performance across all lengths, with minimal degradation. 
Furthermore, the advantages of our approach are even more pronounced in the small-model regime. As shown in Table~\ref{tab:scalable-input-length}, the compact TimeMAR-S model (with fewer than 6M parameters) consistently and significantly outperforms all other small-model baselines across both datasets and all sequence lengths. On the ETTh dataset, TimeMAR-S achieves remarkably low and stable Discriminative and Context-FID scores, demonstrating performance that is not only superior to its direct competitors but also competitive with much larger models like SDformer. While a moderate performance gap exists between TimeMAR-S and TimeMAR-L on the more challenging Energy-256 setting, TimeMAR-S still maintains a substantial lead over the next best small baseline, Diffusion-TS. This underscores the exceptional parameter efficiency and scalability of the TimeMAR architecture, proving that high-quality, long-horizon time series generation is feasible even under stringent model size constraints.

Figure~\ref{fig:fig1} further demonstrates TimeMAR’s efficiency, achieving the lowest FID on Energy—even among methods with similar model size. Figure~\ref{fig:fig2} provides qualitative comparisons, showing that TimeMAR accurately captures both global trend shapes and local seasonal fluctuations. Notably, the decomposed coarse components correspond to fundamental seasonal structures, underscoring the interpretability enabled by our architecture.

\begin{figure*}[h]
    \centering
    \includegraphics[width=0.80\textwidth]{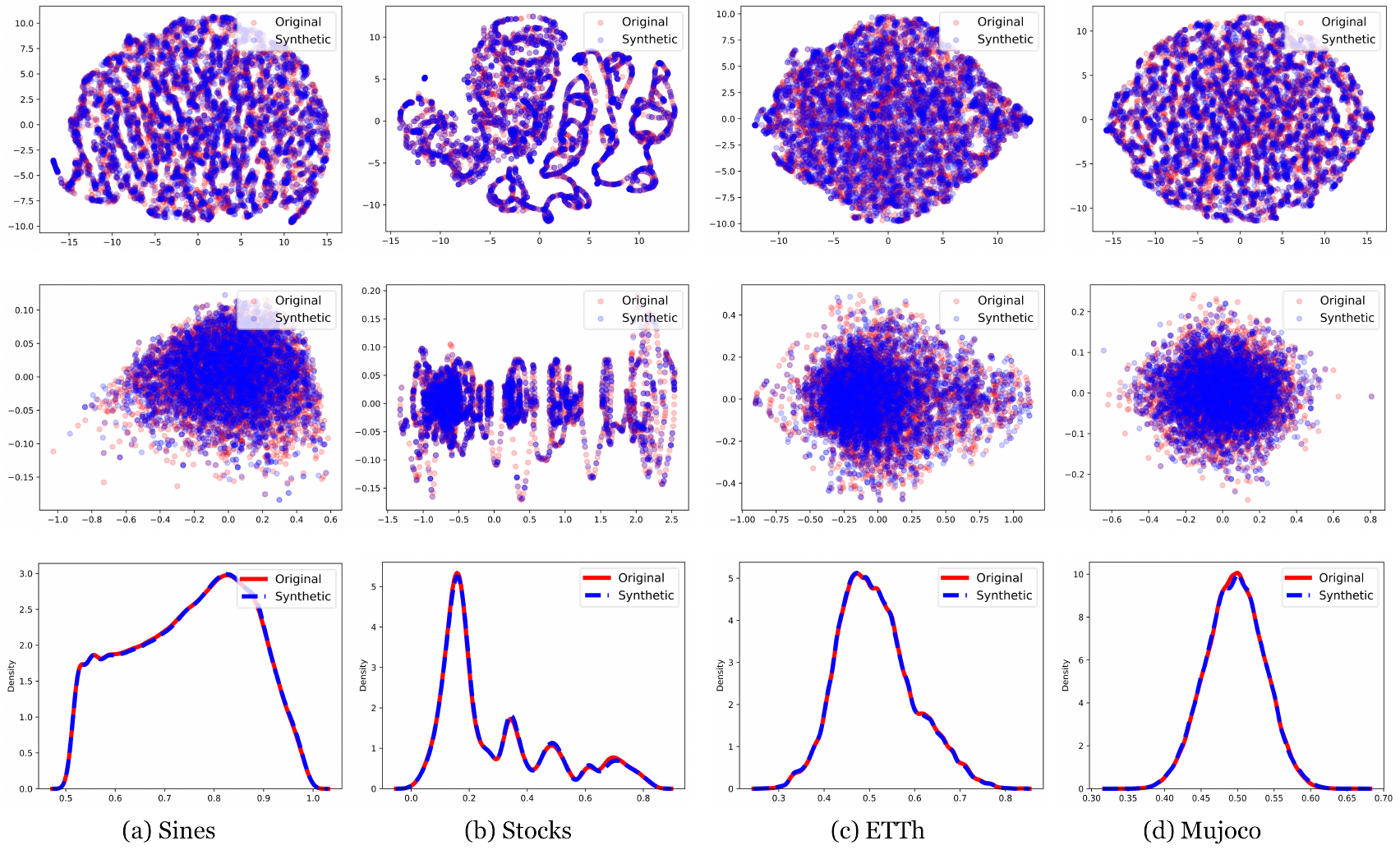}
    \vspace{-0.6em}
    \caption{
       Visualization of synthesized time series by t-SNE (1st row), PCA (2nd row), and Kernel density estimation (3rd row)
    }
    \label{fig:6}
\end{figure*}

\begin{figure}[t]
  \centering
  \begin{minipage}[b]{0.4\textwidth}
    \centering
    \includegraphics[width=\textwidth]{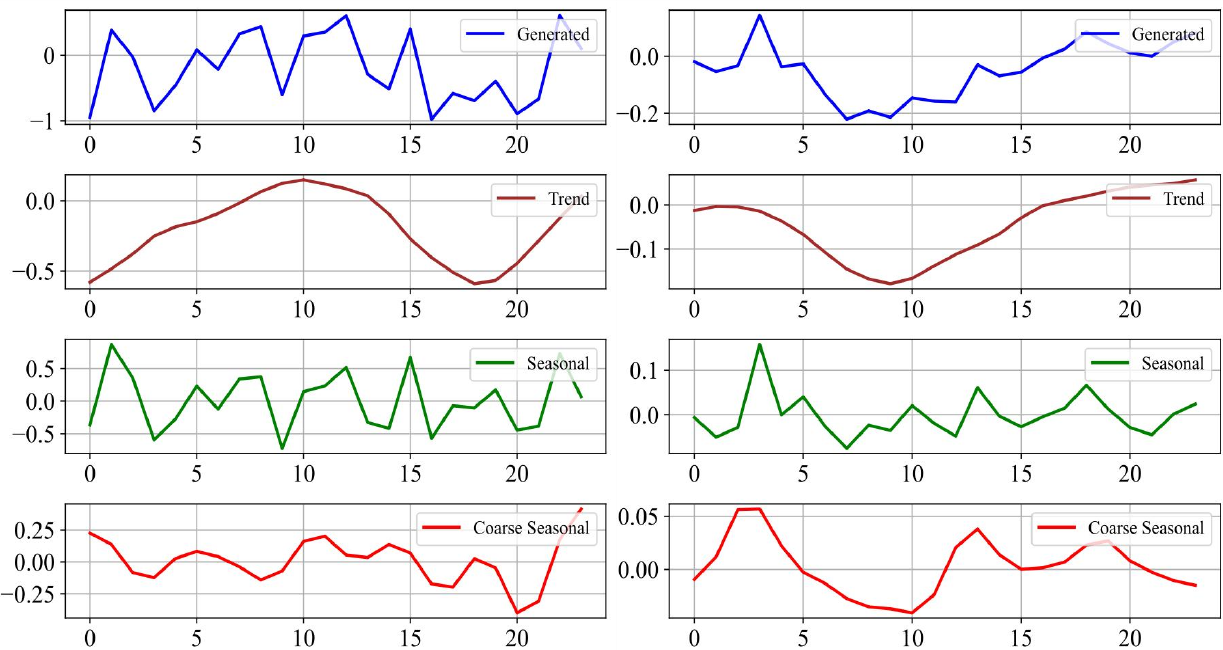}
    \vspace{-2em}
    \caption{Visualization of the Interpretable Generation Process. Left: Energy Dataset; Right: ETTh Dataset.}
    \label{fig:fig2}
  \end{minipage}
\end{figure}

Finally, we present additional visual analysis of TimeMAR. Figure~\ref{fig:6} further complements these results with visualizations based on t-SNE (top row), PCA (middle row), and kernel density estimation (bottom row). The high degree of alignment between the synthetic and real data distributions—evident from both the overlapping embeddings and the closely matched density curves—demonstrates that TimeMAR effectively preserves the statistical and structural properties of original time series.

\begin{table}[tb]
\centering
\setlength{\tabcolsep}{4pt}
\renewcommand{\arraystretch}{1.20}
{\fontsize{8}{9}\selectfont
\begin{tabular}{llccc}
\toprule
\textbf{Metric} & \textbf{Method} & \textbf{Stocks} & \textbf{ETTh} & \textbf{MuJoCo} \\
\midrule
\multirow{5}{*}{\rotatebox{90}{\makecell{\textbf{Discriminative}\\\textbf{Score ↓}}}}
& \textbf{Base}                  & \textbf{0.007$\pm$.003} & \textbf{0.004$\pm$.002} & \textbf{0.005$\pm$.003} \\
& w/o Frequency                  & 0.007$\pm$.005 & 0.005$\pm$.003 & 0.005$\pm$.003 \\
& w/o Decompose                  & 0.010$\pm$.009 & 0.006$\pm$.003 & 0.007$\pm$.003 \\
& w/o Multi-Scale Quant          & 0.010$\pm$.005 & 0.006$\pm$.006 & 0.015$\pm$.005 \\
& w/o Coarse-Guided              & 0.008$\pm$.002 & 0.004$\pm$.002 & 0.005$\pm$.004 \\
\midrule
\multirow{5}{*}{\rotatebox{90}{\makecell{\textbf{Context-FID}\\\textbf{Score ↓}}}}
& \textbf{Base}                  & \textbf{0.001$\pm$.000} & \textbf{0.002$\pm$.000} & \textbf{0.003$\pm$.000} \\
& w/o Frequency                  & 0.004$\pm$.001 & 0.003$\pm$.000 & 0.005$\pm$.000 \\
& w/o Decompose                  & 0.001$\pm$.000 & 0.004$\pm$.000 & 0.004$\pm$.000 \\
& w/o Multi-Scale Quant          & 0.006$\pm$.001 & 0.022$\pm$.001 & 0.021$\pm$.002 \\
& w/o Coarse-Guided              & 0.003$\pm$.001 & 0.003$\pm$.000 & 0.004$\pm$.000 \\
\bottomrule
\end{tabular}}
\caption{Results of ablation study on TimeMAR-S.}
\label{tab:ablation}
\end{table}

\subsection{Ablation Studies}
To assess the impact of each component in our framework, we conduct ablation studies on three representative datasets: \textit{Stocks}, \textit{ETTh}, and \textit{MuJoCo}. As shown in Table~\ref{tab:ablation}, we compare the full model against several variants by individually removing: (1) the frequency-domain cross-attention in the fine seasonal encoder (w/o Frequency), (2) the trend-seasonal decomposition module (w/o Decompose), (3) the hierarchical quantization mechanism (w/o Multi-Scale Quant), and (4) the use of coarse seasonal guidance during fine seasonal reconstruction (w/o Coarse-Guided).

The results clearly demonstrate that all four components play a critical role in ensuring robust performance. In particular, \textbf{w/o Multi-Scale Quant} causes a sharp increase in Context-FID, especially on \textit{ETTh} and \textit{MuJoCo}, which confirms the necessity of hierarchical tokenization for faithfully capturing both global structures and fine-grained temporal variations. \textbf{w/o Decompose} also leads to noticeable performance degradation across both metrics, highlighting the effectiveness of explicit trend-seasonal separation. Furthermore, both \textbf{w/o Frequency} and \textbf{w/o Coarse-Guided} consistently underperform the full model, validating the benefits of frequency-domain interactions and guided reconstruction in refining seasonal patterns. Overall, these results verify that each component meaningfully contributes to the model, and their integration is essential for optimal generation quality.

\section {Conclusion and Future Works}
In this paper, we propose \textbf{TimeMAR}, a multi-scale framework for time series generation. By integrating coarse-to-fine autoregressive modeling, dual-path VQ-VAE encoding, and coarse-guided seasonal reconstruction, our method effectively addresses two key challenges in time series synthesis: hierarchical temporal dependencies and structural heterogeneity. Extensive experiments on both real-world and synthetic datasets demonstrate that TimeMAR consistently outperforms existing generative models across multiple evaluation metrics, achieving high generation fidelity, strong temporal consistency, and robust downstream forecasting performance.

In future work, we aim to develop a unified generative framework capable of handling diverse time series tasks such as unconditional generation, conditional generation conditioned on partial observations or external signals, and missing value imputation, thereby enhancing the flexibility and applicability of our approach in real-world scenarios.

\begin{acks}
This work was partially supported by the National Natural Science Foundation of China (62472174), the ECNU Multifunctional Platform for Innovation (001), and the Fundamental Research Funds for the Central Universities.
\end{acks}

\bibliographystyle{ACM-Reference-Format}
\bibliography{www2026}

\clearpage
\appendix

\section{Detailed Hyperparameters}
This section provides a overview of the hyperparameter configurations employed in the two-stage training pipeline. Table~\ref{tab:stage1} delineates the specific settings used for training the Vector Quantized Variational Autoencoder (VQ-VAE) in Stage 1. The parameters were adjusted based on dataset characteristics to optimize performance. Specifically, a smaller vocab\_size was chosen for the Stocks dataset to suit its smaller data volume. For the fMRI dataset, which has high-dimensional channels, we employed a shallower architecture (with a reduced ch\_mult) and a wider range of patch\_nums to lower the compression ratio, which led to better reconstruction fidelity. Subsequently, Table~\ref{tab:stage2} summarizes the hyperparameters for the autoregressive transformer model trained in Stage 2.
 
\begin{table*}[b]
    \centering
    \begin{tabular}{l|c c c c c c}
        \toprule
        Dataset & vocab\_size & z\_channels & ch & ch\_mult & enc/dec layers & patch\_nums \\
        \midrule
        Sines   & 512  & 256 & 256 & 1, 1, 2 & 3 & 1, 2, 3, 4, 5, 6 \\
        Stocks  & 384  & 256 & 256 & 1, 1, 2 & 3 & 1, 2, 3, 4, 5, 6 \\
        ETTh    & 512  & 256 & 256 & 2, 2, 2 & 3 & 1, 2, 3, 4, 5, 6 \\
        MuJoCo  & 512  & 256 & 256 & 1, 1, 2 & 3 & 1, 2, 3, 4, 5, 6 \\
        Energy  & 512  & 256 & 256 & 1, 1, 2 & 3 & 1, 2, 3, 4, 5, 6 \\
        fMRI    & 1024 & 256 & 256 & 1, 2 & 3 & 1, 2, 3, 4, 6, 8, 10, 12 \\ 
        \bottomrule
    \end{tabular}
    \caption{Hyperparameters of the TimeMAR-L model for Stage 1.  \textit{vocab\_size} is the size of the codebook. \textit{z\_channels} is the number of latent space channels. \textit{ch} is the base channel count. \textit{ch\_mult} are the channel multipliers per resolution stage. \textit{patch\_nums} indicate the number of patches for each scale.}
    \label{tab:stage1}
\end{table*}

\begin{table*}[b]
    \centering
    \begin{tabular}{l|c c c c c}
        \toprule
        Dataset & embed\_dim & layers & attention heads & fc\_rate & Model Size (M) \\
        \midrule
        Sines & 1024 & 1 & 16 & 4 & 61.2\\
        Stocks & 1024 & 1 & 16 & 4 & 60.9\\
        ETTh & 1024 & 1 & 16 & 4 & 82.9\\
        MuJoCo & 1024 & 1 & 16 & 4 & 61.2\\
        Energy & 1024 & 1 & 16 & 4 & 61.3\\
        fMRI & 1024 & 1 & 16 & 4 & 63.8\\
        \bottomrule
    \end{tabular}
    \caption{Hyperparameters of the TimeMAR-L model for Stage 2. \footnotesize \textit{fc\_rate} is the FFN expansion factor.}
    \label{tab:stage2}
\end{table*}


\section{Computational Efficiency}
The primary computational bottleneck during inference stems from the extended autoregressive sequence length, which results in a time complexity of $\mathcal{O}(L^{3})$. Here, $L$ represents the total sequence length, defined as $L = \sum_{k=1}^{K} \text{len}(r_{k})$ where $\{r_{1}, r_{2}, \ldots, r_{K}\}$ denote the $K$ discrete token maps. This cubic complexity arises from the self-attention mechanism in Transformer architectures, making inference time highly sensitive to sequence length scaling. 
While this hierarchical structure inevitably increases the computational load compared to non-hierarchical autoregressive methods, the resulting sampling time remains highly competitive. As evidenced by the comparative results in Table~\ref{tab:appendix_sampling_time}, TimeMAR requires a fraction of the time of the Diffusion-TS baseline while delivering superior forecasting performance (as shown in Table~\ref{tab:evaluation-metrics}), achieving a favorable trade-off between efficiency and generation quality.

\begin{figure}[H]
    \centering
    \includegraphics[width=\columnwidth]{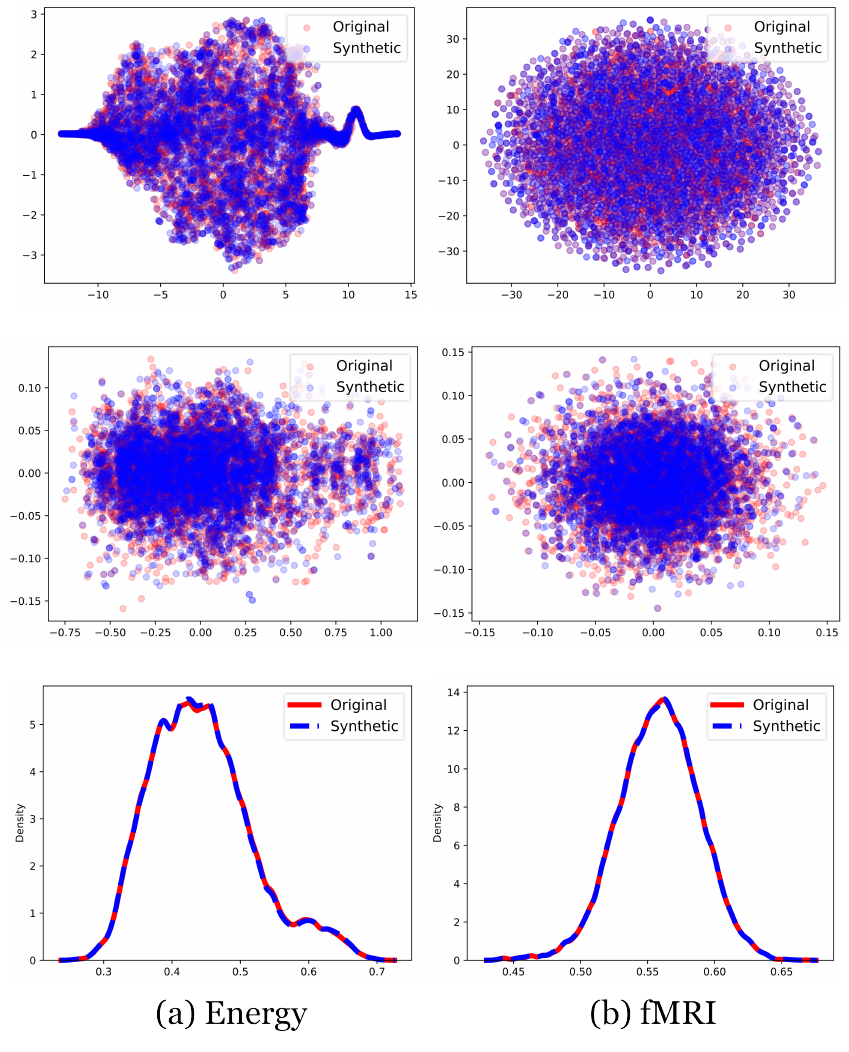} 
    \vspace{-2em}
    \caption{
      Additional visualization of synthesized time series by t-SNE (1st row), PCA (2nd row), and Kernel density estimation (3rd row)
    }
    \label{fig:7}
\end{figure}

\section{Training Detail}

Our primary experiments are executed on an Nvidia 3090 GPU using the AdamW optimizer\cite{adamw}. The training procedure consists of two distinct stages, with training hyperparameters summarized in Table~\ref{tab:train detail}. The learning rate decays by a factor of 0.5 every fixed number of epochs (specified by the decay epoch row).

\begin{table}[H]
    \centering
    \begin{minipage}{0.45\textwidth}
    \centering
    \begin{tabular}{lcc}
        \toprule
        Method & Sampling time (s) \\
        \midrule
        TimeMAR-S & 1.67 \\
        TimeMAR-L & 3.29 \\
        Sdformer & 1.19 \\
        Diffusion-TS & 13.25 \\
        \bottomrule
    \end{tabular}
    \caption{Inference sampling time comparison on the Stock dataset.}
    
    \label{tab:appendix_sampling_time}
    \end{minipage}
    \hfill
    \begin{minipage}{0.45\textwidth}
    \centering
    \vspace{-0.5em}
    \begin{tabular}{l|c c }
        \toprule
        Parameter & stage 1 & stage 2 \\
        \midrule
        epoch  & 5000  & 900 \\
        lr   & 1e-4  & 1e-4 \\
        decay epoch  & 500  & 90 \\
        decay rate    & 0.5  & 0.5 \\
        batch size  & 128  & 512 \\
        \bottomrule
    \end{tabular}
    \caption{Stage 1 and Stage 2 training configurations for TimeMAR-L. \footnotesize \textsuperscript{†} For the \textbf{Energy} dataset, Stage 2 uses a starting learning rate of $5 \times 10^{-5}$.}
    \label{tab:train detail}
    \end{minipage}
\end{table}

\section{Dataset and Metrics}
\subsection{Dataset}
We evaluate on six datasets, configured following the Diffusion‐TS\cite{diffusionts} and SDFormer\cite{sdformer} setups: The Sines dataset\cite{yoon2019time} contains 10,000 samples with 5 dimensions, where each feature is an independent sinusoid generated using distinct frequencies and phases. The Stocks dataset spans Google’s daily stock prices from 2004 to 2019, totaling 3,773 observations across 6 dimensions. The ETTh dataset\cite{zhou2021informer} captures electricity transformer load from July 2016 to July 2018, providing 17,420 samples of 7 variables that reflect real-world power‐system operations. In the Mujoco dataset\cite{mujoco}, 10,000 samples and 14 dimensions record physics-based simulator states used in reinforcement‐learning environments. The Energy dataset comes from UCI’s low-energy building study, offering 19,711 observations and 28 features for appliance-energy–use regression modeling. The fMRI dataset features 10,000 samples and 50 dimensions, simulating realistic blood-oxygen-level-dependent (BOLD) time series to serve as a causal discovery benchmark. 

\begin{table*}[b]
\centering
\setlength{\tabcolsep}{6pt}
\renewcommand{\arraystretch}{1.2}
\begin{tabular}{c|cc|cc|cc}
\toprule
\multirow{2}{*}{\textbf{patch\_num} $\backslash$ \textbf{vocab\_size}} 
 & \multicolumn{2}{c|}{256} 
 & \multicolumn{2}{c|}{384} 
 & \multicolumn{2}{c}{512} \\
\cmidrule{2-7}
 & ds & fid & ds & fid & ds & fid \\
\midrule
1,2,3,6       &0.010$\pm$.006 & 0.001$\pm$.000 & 0.004$\pm$.006 & 0.001$\pm$.000 & 0.012$\pm$.009& 0.001$\pm$.000\\
1,2,3,4,5,6   &0.013$\pm$.004 & 0.001$\pm$.000 & 0.007$\pm$.003 & 0.001$\pm$.000 & 0.007$\pm$.008& 0.001$\pm$.000\\
\bottomrule
\end{tabular}
\caption{Parameter sensitivity analysis of TimeMAR-S on the Stocks dataset. The table reports performance using Discriminative Score (lower is better) and Context-FID Score (lower is better) metrics across different patch number and vocabulary size configurations.}
\end{table*}

\subsection{Metrics}
\textbf{Discriminative Score}. The discriminative score assesses the distinguishability between real and synthetic data. Specifically, a two-layer LSTM classifier is trained to differentiate between the ground truth and the generated sequences. The score is calculated as $|\text{Accuracy} - 0.5|$, reflecting the model's ability to discriminate beyond random chance.

\textbf{Context-FID}. The Contextual Fréchet Inception Distance (Context-FID) quantifies the fidelity of synthetic time series by measuring the statistical distance between representations of ground truth and generated sequences within the same local context. This metric captures both the quality and contextual consistency of the generated data.

\begin{figure}[H] 
\centering
\begin{minipage}{0.48\textwidth}
\begin{algorithm}[H]
\caption{Training Framework for Dual-path VQ-VAE on Time Series}
\label{alg:vqvae}
\begin{algorithmic}[1]
\State \textbf{Input:} Time series $\mathbf{X} \in \mathbb{R}^{T \times D}$ 
\State \textbf{Output:} Trained model $\{E_t, E_s, D_t, D_s, \mathcal{Q}, C\}$

\For{epoch $\gets 1$ to $N$}
    \State \textit{Multi-scale Decomposition:}
    \State $\mathbf{X}_s, \mathbf{X}_t \gets \text{Decomp}(\mathbf{X})$ \Comment{Seasonal/trend separation}
    \State $\mathbf{X}_c \gets \text{MultirateSignalProcessing}(\mathbf{X}_s)$ \Comment{Coarse seasonal separation}
    
    \State \textit{Dual-path Encoding:}
    \State $\mathbf{Z}_t \gets E_t(\mathbf{X}_t),\ \mathbf{Z}_s \gets E_s(\mathbf{X}_s)$ 
    \State $\mathbf{Z}_c \gets E_t(\mathbf{X}_c)$ \Comment{Shared trend encoder}
    
    \State \textit{Feature Fusion:}
    \State $\mathbf{Z}_f \gets \text{Fusion}(\mathbf{Z}_t, \mathbf{Z}_s, \mathbf{Z}_c)$ \Comment{Representation interaction}
    
    \State \textit{Vector Quantization:}
    \State $\hat{\mathbf{Z}}_f, \mathcal{L}_{\text{vq}} \gets \mathcal{Q}(\mathbf{Z}_f)$ \Comment{Codebook projection}
    
    \State \textit{Latent Disentanglement:}
    \State $\hat{\mathbf{Z}}_t, \hat{\mathbf{Z}}_s, \hat{\mathbf{Z}}_c \gets \text{LatentDecomp}(\hat{\mathbf{Z}}_f)$
    
    \State \textit{Dual-path Decoding:}
    \State $\hat{\mathbf{X}}_t \gets D_t(\hat{\mathbf{Z}}_t)$ \Comment{Trend reconstruction}
    \State $\hat{\mathbf{X}}_c \gets D_t(\hat{\mathbf{Z}}_c)$ \Comment{Coarse seasonal guide}
    \State $\hat{\mathbf{X}}_s \gets D_s(\hat{\mathbf{Z}}_s, \hat{\mathbf{X}}_c)$ \Comment{Conditional decoding}
    
    \State \textit{Loss Computation:}
    \State $\mathcal{L}_{\text{total}} \gets \mathcal{L}_{\text{recon}} + \lambda\mathcal{L}_{\text{vq}}$
    
    \State \textit{Parameter Update:}
    \State Backpropagate $\nabla\mathcal{L}_{\text{total}}$ 
    \State Update codebook $C$ and model parameters
\EndFor
\end{algorithmic}
\end{algorithm}
\end{minipage}
\hfill
\begin{minipage}{0.48\textwidth}
\begin{algorithm}[H]
\caption{Multi-Scale Quantization}
\label{alg:quant}
\begin{algorithmic}[1]
\State \textbf{Input:} Fused representation $f$; Number of scales $K$; Resolutions $(l_k)_{k=1}^K$ \\
\textbf{Output:} Multi-scale token sequence $R$
\State $R \gets [\,]$ \Comment{Initialize token list}
\For{$k \gets 1$ to $K$}
    \State $r_k \gets Q(\text{interpolate}(f, l_k))$ \Comment{Quantize feature at resolution $(l_k)$}
    \State $R \gets \text{queue\_push}(R, r_k)$ \Comment{Store multi-scale token}
    \State $z_k \gets \text{lookup}(Z, r_k)$ \Comment{Retrieve embedding}
    \State $z_k \gets \text{interpolate}(z_k, l_k)$ \Comment{Upsample to base resolution}
    \State $f \gets f - \phi_k(z_k)$ \Comment{Residual update}
\EndFor
\State \Return{$R$}
\end{algorithmic}
\end{algorithm}
\end{minipage}
\end{figure}

\section{Algorithms}
Algorithm~\ref{alg:vqvae} outlines the training procedure for the proposed dual-path VQ-VAE framework for time series modeling. Algorithm~\ref{alg:quant} presents the multi-scale quantization strategy used to generate multi-scale discrete token sequence.

\begin{table*}[b]
    \centering
    \scriptsize
    \setlength{\tabcolsep}{4pt}
    \renewcommand{\arraystretch}{1.15}
    {\fontsize{8}{9.5}\selectfont
    \begin{tabular}{llcccccc}
    \toprule
    \textbf{Metric} & \textbf{Method} & \textbf{Sines} & \textbf{Stocks} & \textbf{ETTh} & \textbf{MuJoCo} & \textbf{Energy} & \textbf{fMRI} \\
    \midrule
    
    \multirow{10}{*}{\shortstack[c]{\textbf{Predictive}\\\textbf{Score} $\downarrow$}}
    & \textbf{TimeMAR-L} & \textbf{0.093±.000} & \underline{0.037±.000} & \textbf{0.118±.002} & \textbf{0.007±.000} & \textbf{0.249±.000} & \textbf{0.089±.003} \\
    & SDformer & \textbf{0.093±.000} & \underline{0.037±.000} & \textbf{0.118±.002} & \textbf{0.007±.001} & \textbf{0.249±.000} & \underline{0.091±.002} \\
    & Diffusion-TS & \textbf{0.093±.000} & \textbf{0.036±.000} & \underline{0.119±.002} & \textbf{0.007±.000} & \underline{0.250±.000} & 0.099±.000 \\
    & TimeGAN     & \textbf{0.093±.019} & 0.038±.001 & 0.124±.001 & 0.025±.003 & 0.273±.004 & 0.126±.002 \\
    & TimeVAE     & \textbf{0.093±.000} & 0.039±.000 & 0.126±.004 & 0.012±.002 & 0.292±.000 & 0.113±.003 \\
    & Diffwave    & \textbf{0.093±.000} & 0.047±.000 & 0.130±.001 & 0.013±.000 & 0.251±.000 & 0.101±.000 \\
    & DiffTime    & \textbf{0.093±.000} & 0.038±.001 & 0.121±.004 & \underline{0.010±.001} & 0.252±.000 & 0.100±.000 \\
    & Cot-GAN     & \underline{0.100±.000} & 0.047±.001 & 0.129±.000 & 0.068±.009 & 0.259±.000 & 0.185±.003 \\
    & Original    & 0.094±.001 & 0.036±.001 & 0.121±.005 & 0.007±.001 & 0.250±.003 & 0.090±.001 \\
    \bottomrule
    
    \end{tabular}
    }
    \caption{Predictive Score on 24-length time series across all datasets. Lower values indicate better performance (↓).}
    \label{tab:ps-metrics}
\centering
    \scriptsize
    \setlength{\tabcolsep}{4pt}
    \renewcommand{\arraystretch}{1.15}
    {\fontsize{8}{9.5}\selectfont
    \begin{tabular}{llcccccc}
    \toprule
    \textbf{Metric} & \textbf{Method} & \textbf{ETTh-64} & \textbf{ETTh-128} & \textbf{ETTh-256} & \textbf{Energy-64} & \textbf{Energy-128} & \textbf{Energy-256} \\
    \midrule
    
    \multirow{10}{*}{\shortstack[c]{\textbf{Predictive}\\\textbf{Score} $\downarrow$}} 
    & \textbf{TimeMAR-L} & \textbf{0.111±.003} & \textbf{0.107±.007} & \underline{0.098±.011} & \textbf{0.246±.000} & \textbf{0.244±.001} & \textbf{0.242±.001} \\
    & SDformer & \underline{0.116±.006} & \underline{0.110±.007} & \textbf{0.095±.008} & \underline{0.247±.001} & \textbf{0.244±.000} & \textbf{0.243±.002} \\
    & Diffusion-TS & \underline{0.116±.000} & \underline{0.110±.003} & 0.109±.013 & 0.249±.000 & 0.247±.001 & 0.245±.001 \\
    & TimeGAN     & 0.132±.008 & 0.153±.014 & 0.220±.008 & 0.291±.003 & 0.303±.002 & 0.351±.004 \\
    & TimeVAE     & 0.118±.004 & 0.113±.005 & 0.110±.027 & 0.302±.001 & 0.318±.000 & 0.353±.003 \\
    & Diffwave    & 0.133±.008 & 0.129±.003 & 0.138±.010 & 0.252±.001 & 0.252±.000 & 0.251±.000 \\
    & DiffTime    & 0.118±.004 & 0.120±.008 & 0.118±.003 & 0.252±.000 & 0.251±.000 & 0.251±.000 \\
    & Cot-GAN     & 0.135±.003 & 0.126±.001 & 0.129±.000 & 0.262±.002 & 0.269±.002 & 0.275±.004 \\
    & Original    & 0.114±.006 & 0.108±.005 & 0.106±.010 & 0.245±.002 & 0.243±.000 & 0.243±.000 \\

    \bottomrule
    \end{tabular}
    \caption{Predictive Score on ETTh and Energy datasets under varying input lengths (64/128/256). Lower values indicate better performance (↓).}
    \label{ps-metrics2}}
\end{table*}

\section{Limitations}

The proposed model faces inherent trade-offs between generation quality and computational efficiency. A primary limitation concerns the scaling of model parameters: achieving higher generation quality typically necessitates a substantial increase in model size, which in turn elevates training resource demands and GPU memory footprint. This is manifested in two key design challenges. \textbf{First}, in long-sequence generation, a fundamental trade-off exists between deepening the encoder-decoder stack to improve representational capacity and extending the autoregressive generation horizon to cover longer contexts. While adding layers enhances expressivity, it inflates parameter counts; conversely, lengthening the generation sequence increases per-step computation, prolonging both training and inference times. \textbf{Second}, a related challenge arises in configuring embeddings for high-dimensional data (e.g.,28-channel energy and 50-channel fMRI). Excessively compressing the embedding dimension can degrade output quality, whereas increasing it to mitigate quality loss substantially enlarges the model's parameter scale, complicating deployment on resource-constrained devices. These constraints collectively highlight an inherent tension between model accuracy and operational efficiency.

\section{Predictive Score Results}

Table~\ref{tab:ps-metrics} reports the \textbf{Predictive Score}~\cite{yoon2019time} of our model. TimeMAR-L achieves competitive, state-of-the-art performance, ranking first or second on most datasets.

Note that this metric exhibits limited discriminative power in some cases (e.g., on \textit{Sines}), where scores for top methods are nearly identical. The main analysis therefore focuses on performance differences in terms of \textbf{Discriminative Score} and \textbf{Context-FID Score} for comprehensive evaluation.

\end{document}